\documentclass[11pt]{article}
\usepackage{coling2016}
\usepackage{times}
\usepackage{url}
\usepackage{latexsym}
\usepackage{amsmath,amsfonts,amssymb,amsthm,bm}

\DeclareMathOperator*{\argmin}{argmin}

\title{Lexicons and Minimum Risk Training for Neural Machine Translation: \\ NAIST-CMU at WAT2016}

\author{Graham Neubig$^{\dag*}$ \\
  $^\dag$Nara Institute of Science and Technology, Japan \\
  $^*$Carnegie Mellon University, USA \\
  {\tt gneubig@cs.cmu.edu}}

\date{}

\begin{document}
\maketitle
\begin{abstract}
This year, the Nara Institute of Science and Technology (NAIST)/Carnegie Mellon University (CMU) submission to the Japanese-English translation track of the 2016 Workshop on Asian Translation was based on attentional neural machine translation (NMT) models.
In addition to the standard NMT model, we make a number of improvements, most notably the use of discrete translation lexicons to improve probability estimates, and the use of minimum risk training to optimize the MT system for BLEU score.
As a result, our system achieved the highest translation evaluation scores for the task.
\end{abstract}

\section{Introduction}
\label{sec:intro}

Neural machine translation (NMT; \cite{kalchbrenner13rnntm,sutskever14sequencetosequence}), creation of translation models using neural networks, has quickly achieved state-of-the-art results on a number of translation tasks \cite{luong15iwslt,sennrich16wmt}.
In this paper, we describe NMT systems for the Japanese-English scientific paper translation task of the Workshop on Asian Translation (WAT) 2016 \cite{WAT2016_overview}.

The systems are built using attentional neural networks \cite{bahdanau15alignandtranslate,luong15effectiveattentional}, with a number of improvements (\S\ref{sec:baseline}).
In particular we focus on two.
First, we follow the recent work of \newcite{arthur16emnlp} in incorporating discrete translation lexicons to improve the probability estimates of the neural translation model (\S\ref{sec:lexicon}).
Second, we incorporate minimum-risk training \cite{shen16minrisk} to optimize the parameters of the model to improve translation accuracy (\S\ref{sec:minrisk}).

In experiments (\S\ref{sec:experiments}), we examine the effect of each of these improvements, and find that they both contribute to overall translation accuracy, leading to state-of-the-art results on the Japanese-English translation task.

\section{Baseline Neural Machine Translation Model}
\label{sec:baseline}

Our baseline translation model is the attentional model implemented in the lamtram toolkit \cite{neubig15lamtram}, which is a combination of the models of \newcite{bahdanau15alignandtranslate} and \newcite{luong15effectiveattentional} that we found to be effective.
We describe the model briefly here for completeness, and refer readers to the previous papers for a more complete description.

\subsection{Model Structure}
\label{sec:model}

Our model creates a model of target sentence $E=e^{\lvert E \rvert}_1$ given source sentence $F=f^{\lvert F \rvert}_1$.
These words belong to the source vocabulary $V_f$, and the target vocabulary $V_e$ respectively.
NMT performs this translation by calculating the conditional probability $p_m(e_i | F, e^{i-1}_1)$ of the $i$th target word $e_i$ based on the source $F$ and the preceding target words $e^{i-1}_1$.
This is done by encoding the context $\langle F, e^{i-1}_1 \rangle$ as a fixed-width vector $\bm{\eta}_i$, and calculating the probability as follows:
\begin{equation}
    \label{eq:softmax}
    p_m(e_i| F, e_1^{i-1}) = \text{softmax}(W_{s} \bm{\eta}_i + \bm{b}_{s}),
\end{equation}
where $W_{s}$ and $\bm{b}_{s}$ are respectively weight matrix and bias vector parameters.
The exact variety of the NMT model depends on how we calculate $\bm{\eta}_i$ used as input, and as mentioned above, in this case we use an attentional model.

First, an \textit{encoder} converts the source sentence $F$ into a matrix $R$ where each column represents a single word in the input sentence as a continuous vector.
This representation is generated using a bidirectional encoder
\begin{eqnarray*}
    \overrightarrow{\bm{r}}_j = \text{enc}( \text{embed}(f_j), \overrightarrow{\bm{r}}_{j-1} ) \\
    \overleftarrow{\bm{r}}_j = \text{enc}( \text{embed}(f_j), \overleftarrow{\bm{r}}_{j+1} ).
\end{eqnarray*}
Here the $\text{embed}(\cdot)$ function maps the words into a representation \cite{bengio06nnlm}, and $\text{enc}(\cdot)$ is long short term memory (LSTM) neural network \cite{hochreiter97lstm} with forget gates set to one minus the value of the input gate \cite{greff15lstmsearchspace}.
For the final word in the sentence, we add a sentence-end symbol to the final state of both of these decoders
\begin{eqnarray*}
    \overrightarrow{\bm{r}}_{\lvert F \rvert+1} = \text{enc}( \text{embed}(\langle \text{s} \rangle), \overrightarrow{\bm{r}}_{\lvert F \rvert} ) \\
    \overleftarrow{\bm{r}}_{\lvert F \rvert+1} = \text{enc}( \text{embed}(\langle \text{s} \rangle), \overleftarrow{\bm{r}}_1 ).
\end{eqnarray*}
Finally we concatenate the two vectors $\overrightarrow{\bm{r}}_j$ and $\overleftarrow{\bm{r}}_j$ into a bidirectional representation $\bm{r}_j$
\begin{equation*}
  \bm{r}_j = [\overleftarrow{\bm{r}}_j; \overrightarrow{\bm{r}}_j].
\end{equation*}
These vectors are further concatenated into the matrix $R$ where the $j$th column corresponds to $\bm{r}_j$.

Next, we generate the output one word at a time while referencing this encoded input sentence and tracking progress with a \textit{decoder} LSTM.
The decoder's hidden state $\bm{h}_i$ is a fixed-length continuous vector representing the previous target words $e^{i-1}_1$, initialized as $\bm{h}_0 = \bm{r}_{\lvert F \rvert + 1}$.
This is used to calculate a context vector $\bm{c}_i$ that is used to summarize the source attentional context used in choosing target word $e_i$, and initialized as $\bm{c}_0 = \bm{0}$.

First, we update the hidden state to $\bm{h}_i$ based on the word representation and context vectors from the previous target time step
\begin{equation}
    \bm{h}_i = \text{enc}( [\text{embed}(e_{i-1}); \bm{c}_{i-1}], \bm{h}_{i-1} ).
\end{equation}

Based on this $\bm{h}_i$, we calculate a similarity vector $\bm{\alpha}_i$, with each element equal to
\begin{equation}
\alpha_{i,j} = \text{sim}(\bm{h}_i,\bm{r}_j).
\end{equation}
$\text{sim}(\cdot)$ can be an arbitrary similarity function.
In our systems, we test two similarity functions, the dot product \cite{luong15effectiveattentional}
\begin{equation}
\text{sim}(\bm{h}_i,\bm{r}_j) := \bm{h}_i^\intercal \bm{r}_j
\end{equation}
and the multi-layered perceptron \cite{bahdanau15alignandtranslate}
\begin{equation}
\text{sim}(\bm{h}_i,\bm{r}_j) := \bm{w}_{a2}^\intercal \text{tanh}(W_{a1}[\bm{h}_i; \bm{r}_j]),
\end{equation}
where $W_{a1}$ and $\bm{w}_{a2}$ are the weight matrix and vector of the first and second layers of the MLP respectively.

We then normalize this into an \textit{attention} vector, which weights the amount of focus that we put on each word in the source sentence
\begin{equation}
\label{eq:attention}
\bm{a}_i = \text{softmax}(\bm{\alpha}_i).
\end{equation}
This attention vector is then used to weight the encoded representation $R$ to create a context vector $\bm{c}_i$ for the current time step
\begin{equation*}
    \bm{c}_i = R \bm{a}_i.
\end{equation*}

Finally, we create $\bm{\eta}_i$ by concatenating the previous hidden state with the context vector, and performing an affine transform
\begin{equation*}
    \bm{\eta}_i = W_{\eta}[\bm{h}_i; \bm{c}_i] + b_{\eta},
\end{equation*}

Once we have this representation of the current state, we can calculate $p_m(e_i | F, e^{i-1}_1)$ according to Equation (\ref{eq:softmax}).
The next word $e_i$ is chosen according to this probability.

\subsection{Parameter Optimization}
\label{sec:optimization}

If we define all the parameters in this model as $\theta$, we can then train the model by minimizing the negative log-likelihood of the training data
\begin{equation*}
    \hat{\theta} = \argmin_{\theta} \sum_{\langle F,~E \rangle} \sum_{i} -\log(p_m(e_i | F, e^{i-1}_1; \theta)).
\end{equation*}

Specifically, we use the ADAM optimizer \cite{kingma14adam}, with an initial learning rate of 0.001.
Minibatches of 2048 words are created by sorting sentences in descending order of length and grouping sentences sequentially, adding sentences to the minibatch until the next sentence would cause the minibatch size to exceed 2048 words.%
\footnote{It should be noted that it is more common to create minibatches with a fixed number of sentences. We use words here because the amount of memory used in processing a minibatch is more closely related to the number of words in the minibatch than the number of sentences, and thus fixing the size of the minibatch based on the number of words leads to more stable memory usage between minibatches.}
Gradients are clipped so their norm does not exceed 5.

Training is allowed to run, checking the likelihood of the development set periodically (every 250k sentences processed), and the model that achieves the best likelihood on the development set is saved.
Once no improvements on the development set have been observed for 2M training sentences, training is stopped, and re-started using the previously saved model with a halved learning rate 0.0005.
Once training converges for a learning rate of 0.0005, the same procedure is performed with a learning rate of 0.00025, resulting in the final model.

\subsection{Search}
\label{sec:search}

At test time, to find the best-scoring translation, we perform beam search with a beam size of 5.
At each step of beam search, the best-scoring hypothesis remaining in the beam that ends with the sentence-end symbol is saved.
At the point where the highest-scoring hypothesis in the beam has a probability less than or equal to the best sentence-ending hypothesis, search is terminated and the best sentence-ending hypothesis is output as the translation. 

In addition, because NMT models often tend to be biased towards shorter sentences, we add an optional ``word penalty'' $\lambda$, where each hypothesis's probability is multiplied by $e^{\lambda|E'|}$ for comparison with other hypotheses of different lengths.
This is equivalent to adding an exponential prior probability on the length of output sentences, and if $\lambda>0$, then this will encourage the decoder to find longer hypotheses.

\section{Incorporating Discrete Lexicons}
\label{sec:lexicon}

The first modification that we make to the base model is incorporating discrete lexicons to improve translation probabilities, according to the method of \newcite{arthur16emnlp}.
The motivation behind this method is twofold:
\begin{description}
\item[Handling low-frequency words:] Neural machine translation systems tend to have trouble translating low-frequency words \cite{sutskever14sequencetosequence}, so incorporating translation lexicons with good coverage of content words could improve translation accuracy of these words.
\item[Training speed:] Training the alignments needed for discrete lexicons can be done efficiently \cite{dyer13fastalign}, and by seeding the neural MT system with these efficiently trained alignments it is easier to learn models that achieve good results more quickly.
\end{description}

The model starts with lexical translation probabilities $p_l(e | f)$ for individual words, which have been obtained through traditional word alignment methods.
These probabilities must first be converted to a form that can be used together with $p_m(e_i | e_1^{i-1}, F)$.
Given input sentence $F$, we can construct a matrix in which each column corresponds to a word in the input sentence, each row corresponds to a word in the $V_E$, and the entry corresponds to the appropriate lexical probability:
\small
\begin{equation*}
    L_F = 
        \begin{bmatrix}
            ~p_{l}(e=1 | f_1)~ & \cdots & p_{l}(e=1|f_{\lvert F \rvert})\\
            \vdots & \ddots & \vdots \\
            ~p_{l}(e={\lvert V_e \lvert} | f_1) & \cdots & p_{l}(e={\lvert V_e \lvert}| f_{\lvert F \rvert})~
        \end{bmatrix}.
\end{equation*}
\normalsize
This matrix can be precomputed during the encoding stage because it only requires information about the source sentence $F$.

Next we convert this matrix into a predictive probability over the next word: $p_l(e_i|F, e^{i-1}_1)$.
To do so we use the alignment probability $\bm{a}$ from Equation (\ref{eq:attention}) to weight each column of the $L_F$ matrix:
\small
\begin{equation*}
    p_{l}(e_i|F, e^{i-1}_1) = L_F \bm{a}_i  =
        \begin{bmatrix}
            p_{l}(e=1|f_1) & \cdots & p_{lex}(e=1|f_{\lvert F \rvert}) \\
            \vdots   & \ddots & \vdots \\
            p_{l}(e={V_e}|f_1) & \cdots & p_{lex}(e={V_e}|f_{\lvert F \rvert})
        \end{bmatrix}
        \begin{bmatrix}
            a_{i,1} \\
            \vdots \\
            a_{i,\lvert F \rvert}
        \end{bmatrix}.
\end{equation*}
\normalsize
This calculation is similar to the way how attentional models calculate the context vector $\bm{c}_i$, but over a vector representing the probabilities of the target vocabulary, instead of the distributed representations of the source words.

After calculating the lexicon predictive probability $p_l(e_i | e_1^{i-1}, F)$, next we need to integrate this probability with the NMT model probability $p_m(e_i | e_1^{i-1}, F)$.
Specifically, we use $p_l(\cdot)$ to bias the probability distribution calculated by the vanilla NMT model by adding a small constant $\epsilon$ to $p_{l}(\cdot)$, taking the logarithm, and adding this adjusted log probability to the input of the softmax as follows:
\begin{equation*}
    p_b(e_i| F, e_1^{i-1}) = \text{softmax}(W_{s}\bm{\eta}_i + b_{s} + \log(p_{l}(e_i | F, e^{i-1}_1) + \epsilon)).
\end{equation*}
We take the logarithm of $p_{l}(\cdot)$ so that the values will still be in the probability domain after the softmax is calculated, and add the hyper-parameter $\epsilon$ to prevent zero probabilities from becoming $-\infty$ after taking the log.
We test various values including $\epsilon=\{10^{-4},10^{-5},10^{-6}\}$ in experiments.

\section{Minimum Risk Training}
\label{sec:minrisk}

The second improvement that we make to our model is the use of minimum risk training.
As mentioned in Section \ref{sec:optimization} our baseline model optimizes the model parameters according to maximize the likelihood of the training data.
However, there is a disconnect between the evaluation of our systems using translation accuracy (such as BLEU \cite{papineni02bleu}) and this maximum likelihood objective.

To remove this disconnect, we use the method of \newcite{shen16minrisk} to optimize our systems directly using BLEU score.
Specifically, we define the following loss function over the model parameters $\theta$ for a single training sentence pair $\langle F,E \rangle$
\begin{equation*}
\mathcal{L}_{F,E}(\theta) = \sum_{E'} \text{err}(E,E') P(E'|F;\theta),
\end{equation*}
which is summed over all potential translations $E'$ in the target language.
Here $\text{err}(\cdot)$ can be an arbitrary error function, which we define as $1-\text{SBLEU}(E,E')$, where $\text{SBLEU}(\cdot)$ is the smoothed BLEU score (BLEU+1) proposed by \newcite{lin04orange}.
As the number of target-language translations $E'$ is infinite, the sum above is intractable, so we approximate the sum by randomly sampling a subset of translations $\mathcal{S}$ according to $P(E|F;\theta)$, then enumerating over this sample:%
\footnote{The actual procedure for obtaining a sample consists of calculating the probability of the first word $P(e_1|F)$, sampling the first word from this multinomial, and then repeating for each following word until the end of sentence symbol is sampled.}
\begin{equation*}
\mathcal{L}_{F,E}(\theta) = \sum_{E' \in \mathcal{S}} \text{err}(E,E') \frac{P(E'|F;\theta)}{\sum_{E'' \in \mathcal{S}} P(E''|F;\theta)}.
\end{equation*}
This objective function is then modified by introducing a scaling factor $\alpha$, which makes it possible to adjust the smoothness of the distribution being optimized, which in turn results in adjusting the strength with which the model will try to push good translations to have high probabilities.
\begin{equation*}
\mathcal{L}_{F,E}(\theta) = \sum_{E' \in \mathcal{S}} \text{err}(E,E') \frac{P(E'|F;\theta)^{\alpha}}{\sum_{E'' \in \mathcal{S}} P(E''|F;\theta)^{\alpha}}.
\end{equation*}

In this work, we set $\alpha=0.005$ following the original paper, and set the number of samples to be 20.

\section{Experiments}
\label{sec:experiments}

\subsection{Experimental Setup}

To create data to train the model, we use the top 2M sentences of the ASPEC Japanese-English training corpus \cite{ASPEC} provided by the task.
The Japanese size of the corpus is tokenized using KyTea \cite{neubig11aclshort}, and the English side is tokenized with the tokenizer provided with the Travatar toolkit \cite{neubig13travatar}.
Japanese is further normalized so all full-width roman characters and digits are normalized to half-width. 
The words are further broken into subword units using joint byte pair encoding \cite{sennrich16bpe} with 100,000 merge operations.

\subsection{Experimental Results}

In Figure \ref{tab:overall} we show results for various settings regarding attention, the use of lexicons, training criterion, and word penalty.
In addition, we calculate the ensemble of 6 models, where the average probability assigned by each of the models is used to determine the probability of the next word at test time.

\begin{table*}[t]
\begin{center}
\begin{tabular}{lll||rrr|rrr|rrr}
    &        &                              & \multicolumn{3}{|c}{ML ($\lambda$=0.0)} & \multicolumn{3}{|c}{ML ($\lambda$=0.8)} & \multicolumn{3}{|c}{MR ($\lambda$=0.0)} \\
    & Attent & Lex ($\epsilon$)             & B    & R     & Rat. & B    & R     & Rat.  & B    & R     & Rat. \\ \hline \hline
(1) & dot    & No                           & 22.9 & 74.4  & 89.9 & 24.7 & 74.3  & 100.9 & 25.7 & 75.4  & 97.3 \\ 
(2) & dot    & Yes ($10^{-4}$)                   & 23.0 & 74.6  & 91.0 & 24.5 & 74.2  & 100.4 & 25.3 & 75.3  & 99.2 \\ 
(3) & dot    & Yes ($10^{-5}$)                   & 23.8 & 74.6  & 91.4 & 25.1 & 74.2  & 100.4 & 25.9 & 75.5  & 98.0 \\ 
(4) & dot    & Yes ($10^{-6}$)                   & 23.7 & 74.4  & 92.1 & 25.3 & 74.3  & 99.6  & 26.2 & 76.0  & 98.6 \\ 
(5) & MLP    & Yes ($10^{-4}$)                   & 23.7 & 75.3  & 88.5 & 25.5 & 75.2  & 97.9  & 26.9 & 76.3  & 98.8 \\ 
(6) & MLP    & Yes ($10^{-5}$)                   & 23.7 & 75.1  & 90.5 & 25.3 & 74.8  & 98.6  & 26.4 & 75.9  & 97.7 \\ 
(7) & MLP    & Yes ($10^{-6}$)                   & 23.9 & 74.6  & 89.4 & 25.8 & 74.6  & 99.3  & 26.3 & 75.7  & 97.3 \\ \hline
(8) & \multicolumn{2}{l||}{(2)-(7) Ensemble} & \multicolumn{3}{|c|}{-}  & 27.3 & 75.8  & 99.8  & 29.3 & 77.3  & 97.9 \\
\end{tabular}
\end{center}
\caption{\label{tab:overall} Overall BLEU, RIBES, and length ratio for systems with various types of attention (dot product or multi-layer perceptron), lexicon (yes/no and which value of $\lambda$), training algorithm (maximum likelihood or minimum risk), and word penalty value.}
\end{table*}

From the results in the table, we can glean a number of observations.
\begin{description}
\item[Use of Lexicons:] Comparing (1) with (2-4), we can see that in general, using lexicons tends to provide a benefit, particularly when the $\epsilon$ parameter is set to a small value.
\item[Type of Attention:] Comparing (2-4) with (5-7) we can see that on average, multi-layer perceptron attention was more effective than using the dot product.
\item[Use of Word Penalties:] Comparing the first and second columns of results, there is a large increase in accuracy across the board when using a word penalty, demonstrating that this is an easy way to remedy the length of NMT results.
\item[Minimum Risk Training:] Looking at the third column, we can see that there is an additional increase in accuracy from minimum risk training. In addition, we can see that after minimum risk, the model produces hypotheses that are more-or-less appropriate length without using a word penalty, an additional benefit.
\item[Ensemble:] As widely reported in previous work, ensembling together multiple models greatly improved performance.
\end{description}

\subsection{Manual Evaluation Results}

The maximum-likelihood trained ensemble system with a word penalty of 0.8 (the bottom middle system in Table \ref{tab:overall}) was submitted for manual evaluation.
The system was evaluated according to the official WAT ``HUMAN'' metric \cite{WAT2016_overview}, which consists of pairwise comparisons with a baseline phrase-based system, where the evaluated system receives +1 for every win, -1 for every tie, 0 for every loss, these values are averaged over all evaluated sentences, then the value is multiplied by 100.
This system achieved a manual evaluation score of 47.50, which was slightly higher than other systems participating in the task.
In addition, while the full results of the minimum-risk-based ensemble were not ready in time for the manual evaluation stage, a preliminary system ensembling the minimum-risk-trained versions of the first four systems (1)-(4) in Table \ref{tab:overall} was also evaluated (its BLEU/RIBES scores were comparable to the fully ensembled ML-trained system), and received a score of 48.25, the best in the task, albeit by a small margin.

\section{Conclusion}
\label{sec:conclusion}

In this paper, we described the NAIST-CMU system for the Japanese-English task at WAT, which achieved the most accurate results on this language pair.
In particular, incorporating discrete translation lexicons and minimum risk training were found to be useful in achieving these results.

\section*{Acknowledgments:} This work was supported by JSPS KAKENHI Grant Number 25730136.

\bibliographystyle{acl}
\bibliography{myabbrv,main}

\end{document}